\documentclass[sigconf]{acmart}

\renewcommand\footnotetextcopyrightpermission[1]{} 
\AtBeginDocument{%
  \providecommand\BibTeX{{%
    \normalfont B\kern-0.5em{\scshape i\kern-0.25em b}\kern-0.8em\TeX}}}

\setcopyright{acmcopyright}
\copyrightyear{2018}
\acmYear{2018}
\acmDOI{XXXXXXX.XXXXXXX}

\acmConference[Conference acronym 'XX]{Make sure to enter the correct
  conference title from your rights confirmation emai}{June 03--05,
  2018}{Woodstock, NY}
\acmPrice{15.00}
\acmISBN{978-1-4503-XXXX-X/18/06}

\begin{document}


\title{FRIH: Fine-grained Region-aware Image Harmonization}

\author{Jinlong Peng$^{1}$, Zekun Luo$^{1}$, Liang Liu$^{1}$, Boshen Zhang$^{1}$, Tao Wang$^{2}$, \\
Yabiao Wang$^{1}$, Ying Tai$^{1}$, Chengjie Wang$^{1}$, Weiyao Lin$^{2}$}

\affiliation{%
\institution{$^1$Tencent Youtu Lab \\
$^2$Shanghai Jiao Tong University}
}

\email{{jeromepeng, zekunluo, leoneliu, boshenzhang, caseywang, yingtai, jasoncjwang}@tencent.com}

\email{{wang_tao1111, wylin}@sjtu.edu.cn}



\begin{abstract}
  Image harmonization aims to generate a more realistic appearance of foreground and background for a composite image. Existing methods perform the same harmonization process for the whole foreground. However, the implanted foreground always contains different appearance patterns. All the existing solutions ignore the difference of each color block and losing some specific details. Therefore, we propose a novel global-local two stages framework for Fine-grained Region-aware Image Harmonization (FRIH), which is trained end-to-end. In the first stage, the whole input foreground mask is used to make a global coarse-grained harmonization. In the second stage, we adaptively cluster the input foreground mask into several submasks by the corresponding pixel RGB values in the composite image. Each submask and the coarsely adjusted image are concatenated respectively and fed into a lightweight cascaded module, adjusting the global harmonization performance according to the region-aware local feature. Moreover, we further designed a fusion prediction module by fusing features from all the cascaded decoder layers together to generate the final result, which could utilize the different degrees of harmonization results comprehensively. Without bells and whistles, our FRIH algorithm achieves the best performance on iHarmony4 dataset (PSNR is 38.19 dB)  with a lightweight model. The parameters for our model are only 11.98 M, far below the existing methods.
\end{abstract}




\maketitle

\section{Introduction}
Image composition plays an essential role in image editing and generation \cite{liu2020composition,azadi2020compositional,qiu2020semanticadv,cheng2020sequential,wang2020image,guo2019progressive,gregor2015draw,van2016conditional}. However, since the source of the implanted foreground object and the new background image are different, it is easy to cause an unrealistic perception of the composite image. Image harmonization is an important operation to address this issue, which aims to make the implanted foreground compatible with the background.

\begin{figure}[!t]
\vspace{1em}
\centering
\includegraphics[width=\columnwidth]{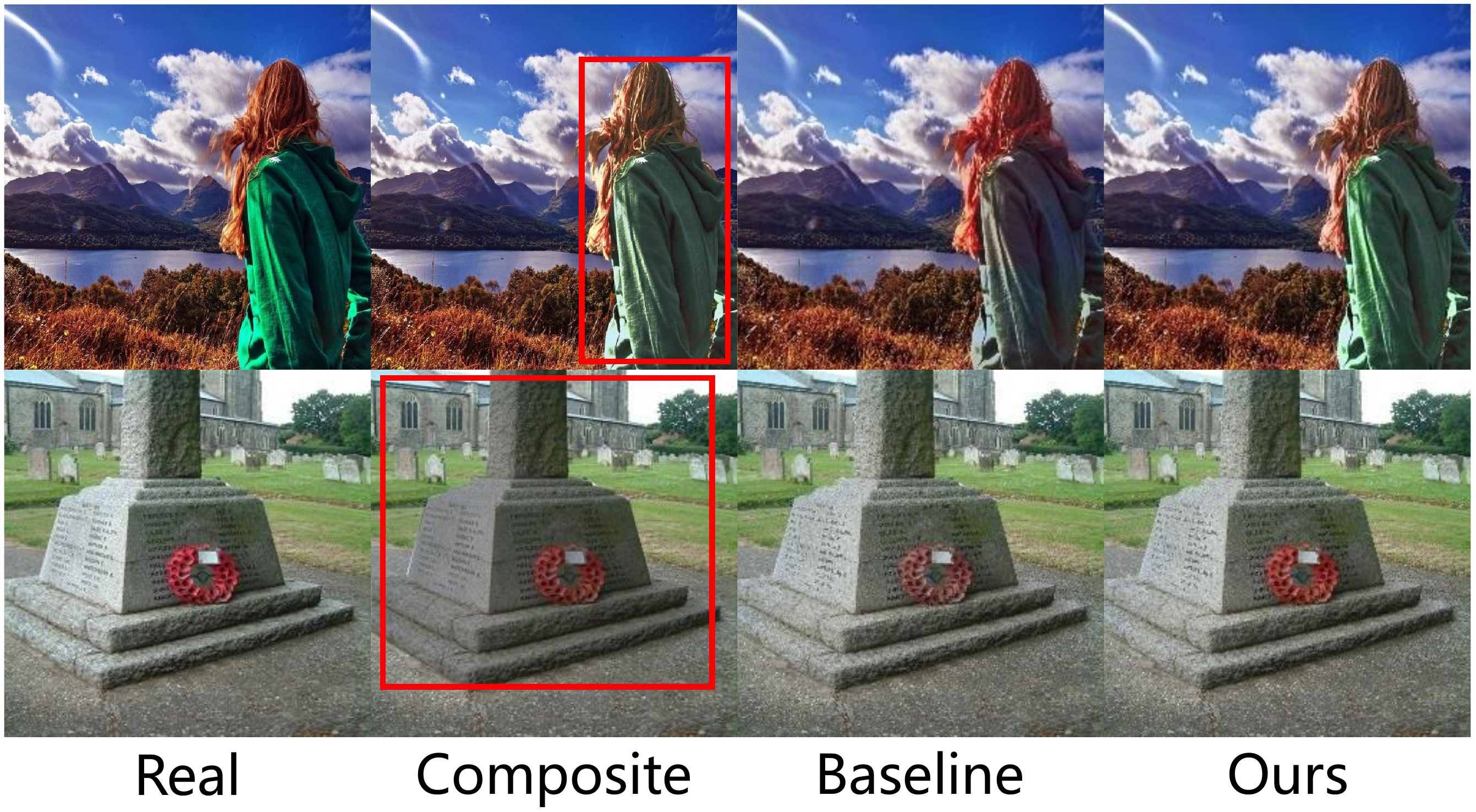}
\caption{Case study on the coarse-grained harmonization problem of different foreground appearance patterns. On the top line, the implanted foreground is a person with green shirt and orange hair. The background contains more orange pixels, leading to a better harmony of the orange hair, while the green saturation in the clothes decreases in baseline. On the bottom line, the implanted foreground is a gray sculpture and a red wreath. The background is brighter, resulting in a brighter overall harmony process in baseline. The gray sculpture looks good while the red wreath becomes too whiter. Clearly, our fine-grained region-aware image harmonization algorithm could solve the above problem well.}
\label{fig:fig1}
\end{figure}

Traditional image harmonization methods mainly focused on low-level feature statistics, such as color distribution matching \cite{pitie2005n,cohen2006color,pitie2007linear}, gradient-domain compositing \cite{perez2003poisson,jia2006drag,tao2013error} and hybrid feature transferring \cite{sunkavalli2010multi}. These methods limit the harmonization performance due to the lack of high level information. There are also several unsupervised or self-supervised image harmonization methods \cite{chen2019toward,zhan2020adversarial,jiang2021ssh}, which does not rely on manual annotation. But they only work better in specific scenes, such as the portraits harmonization, lack of universality. Recently, the supervised encoder-decoder harmonization methods \cite{tsai2017deep,cong2020dovenet,ling2021region,guo2021intrinsic} have achieved superior performance based on the construction of the large training datasets.

However, there is still a major problem of existing solutions. The implanted foreground always contains different appearance patterns. The similarity between the composite image and the real ground-truth image varies among different sub-regions. Some sub-regions in the composite image are relatively similar to the real ground-truth image, while others may be much far away from the target. Thus different sub-regions need different harmonization operations. Existing methods perform the same harmonization process for the whole foreground, ignoring the difference of each color block and losing some specific details. As can be seen in Figure \ref{fig:fig1}, the foreground of the first case is a person with green shirt and orange hair. The foreground of the second case is a gray sculpture and a red wreath. If applying the coarse-grained harmony process (baseline), only part of the foreground (the orange hair and the gray sculpture) has satisfactory harmonization performance, while the other part (the green shirt and the red wreath) is ineffective.


In order to solve the above problem, we propose a simple, novel and effective global-local framework for Fine-grained Region-aware Image Harmonization (FRIH), which is a two-stage network. The first stage includes the base network (a simple U-Net \cite{ronneberger2015u} alike network), in which the composite image and the whole input foreground mask is used to make a global coarse-grained harmonization. In the second stage, we 
adjust the global harmonization performance according to the region-aware local feature. Specifically, we adaptively cluster the input foreground mask into several submasks by the corresponding pixel RGB values in the composite image. The number of the submasks is adaptive for different input images. Each submask and the coarsely adjusted image are concatenated respectively and fed into a lightweight cascaded module. Moreover, in order to ultilize the diffenrent levels of the harmonization results comprehensively, we further designed a fusion prediction module by fusing features from all the cascaded decoder layers together to generate the final result.
Note that our two-stage network FRIH is trained end-to-end. If the two stages are trained separately, the submasks information can not affect the training of the first stage, not able to maximize the optimization of
the overall performance.

Our FRIH method achieves the best performance in iHarmony4 \cite{cong2020dovenet} dataset with a lightweight model. The PSNR of our method is 38.19 dB, performing better than all the existing SOTA methods \cite{cong2020dovenet,guo2021intrinsic,ling2021region}. Specifically, the sub-regions details of the two cases in Figure \ref{fig:fig1} are both handled well by our method. Besides, our model has only 11.98 M parameters, which is only 1/5 of the existing SOTA methods \cite{cong2020dovenet,guo2021intrinsic,ling2021region}. The base network has 9.30 M parameters and the lightweight cascaded module together with the embedded fusion prediction module has 2.68 M parameters, only accounting for 22.4\% of the parameters of the whole FRIH network. The main contributions of this paper are listed as follows:




1. We propose a simple, novel and effective global-local framework for fine-grained region-aware image harmonization (the first as far as we know). The local submasks are generated adaptively to adjust the global coarse-grained harmonization.

2. We design a lightweight cascaded module to integrate the global coarsely adjusted image and the region-aware local feature, refining the harmonization performance. And the fusion prediction module is further proposed to utilize the different degrees of harmonization results comprehensively.

3. Our FRIH achieves the best PSNR (38.19) on iHarmony4 dataset. Besides, the parameters of our model are only 11.98 M, which is far lower than existing methods.

\section{Related Works}

\subsection{Statistics-based Harmonization Methods}

Traditional image harmonization methods mainly focused on low-level feature statistics, such as color distribution matching \cite{pitie2005n,cohen2006color,pitie2007linear,song2020illumination}, gradient-domain compositing \cite{perez2003poisson,jia2006drag,tao2013error} and hybrid feature transferring \cite{sunkavalli2010multi}. These methods did not consider the realism of the composite images. Several other methods \cite{lalonde2007using,xue2012understanding,zhu2015learning} further applied high-level image feature to design the visual reality assessment mechanism. Despite the better optimization standards, the basis of these methods is still statistics methods, limiting the harmonization performance. While our FRIH follows the currently mainstream supervised encoder-decoder framework, which has the potential for better harmonization results.

\subsection{Encoder-decoder Harmonization Methods}

Comparing with traditional statistics-based methods, recent encoder-decoder harmonization methods have achieved superior performance. The pioneering end-to-end CNN method DIH \cite{tsai2017deep} encoded the input image and foreground mask, which was then decoded to the harmonized image and scene parsing image. Based on the encoder-decoder framework, several methods \cite{cun2020improving,cong2020dovenet,hao2020image,cong2021bargainnet} applied the attention mechanism to learn foreground and background appearance feature separately for harmonization. Moreover, RainNet \cite{ling2021region} designed a region-aware adaptive instance normalization module to transfer the visual style from background to foreground. IIH \cite{guo2021intrinsic} proposed intrinsic image harmonization framework by disentangling the composite image into reflectance and illumination for further separate harmonization. Guo \textit{et. al} \cite{guo2021image} integrated the transformer structure \cite{vaswani2017attention,dosovitskiy2021image} to the encoder-decoder harmonization network. However, all these methods only divided the foreground and background as different regions. They did not make a finer region division inside the foreground. Differently, our FRIH is the first method to propose the fine-grained region-aware harmonization framework, which could get more precise harmonization results.


\subsection{Cascade Mechanism}

Cascade mechanism is a direct but useful strategy in many computer vision tasks \cite{cai2018cascade,chen2019hybrid,peng2020chained}. The key idea is to use an extra cascaded module to refine the results from the previous stages. In object detection task, Cascade R-CNN \cite{cai2018cascade} used a sequence of detectors trained with increasing IoU thresholds to gradually refine the detection results from the previous detectors. CU-Net \cite{liu2019cu} connected two U-Net networks and used two stage loss supervision for more accurate segmentation. Cascade EF-GAN\cite{wu2020cascade} decomposed the expression editing task into three steps and applied three cascaded GANs to solve each of them. Most of these methods used the same architecture of base network to design their cascaded module, leading to a large increase of model size. Different from those methods, our cascaded module is lightweight and 
effective. The lightweight module together with the embedded fusion prediction module only accounts for 22.4\% of the parameters of the whole FRIH network.

\begin{figure*}
\centering
\includegraphics[width=0.98\textwidth]{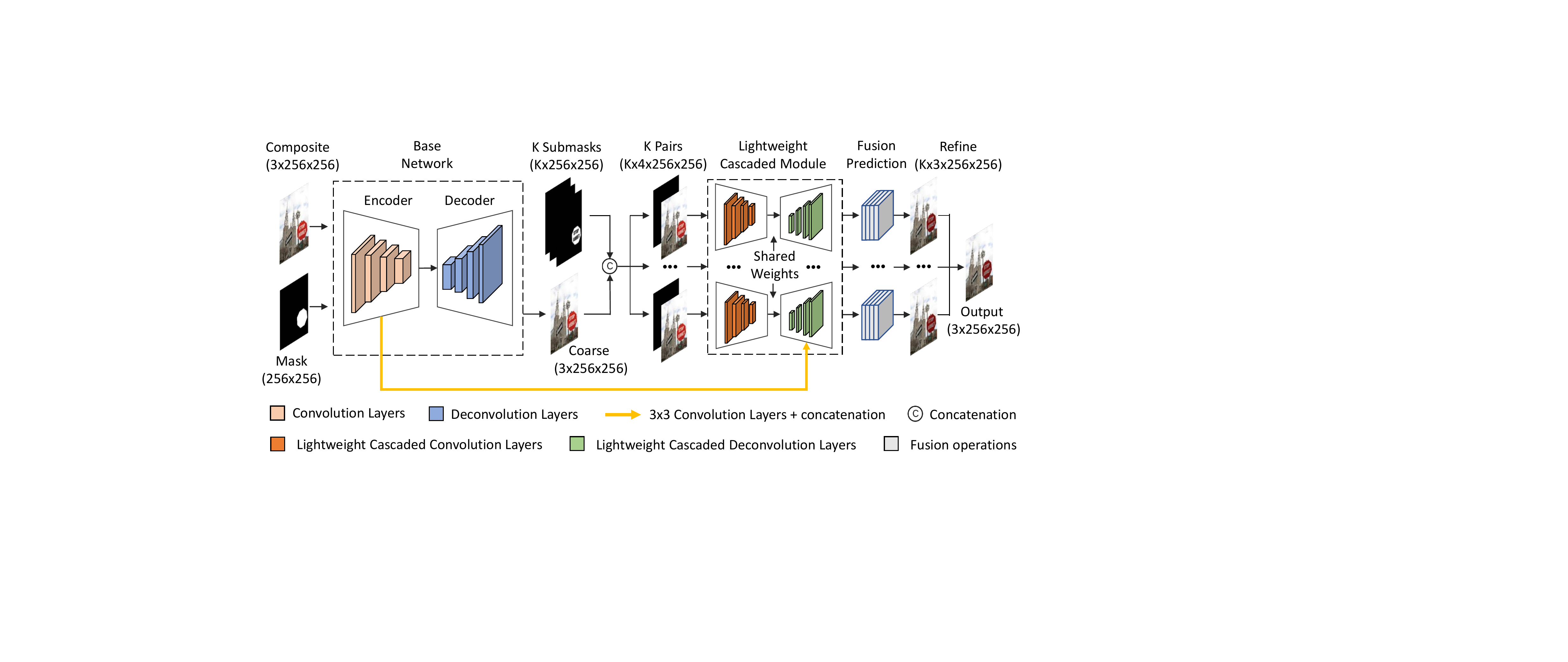}
\vspace{1em}
\caption{Our proposed FRIH framework. In the first stage, we feed the composite image and the foreground mask into the base network to obtain the coarsely adjusted image. In the second stage, the coarsely adjusted image and the extracted submasks are concatenated and fed into the lightweight cascaded module together with the embedded fusion prediction module, which adjusts the sub-regions and generates final refined harmonious images. Note that to keep the figure clean, we omit the skip connections between the encoder and the decoder, both in the base network and the lightweight cascaded module.}
\label{fig:fig2}
\vspace{1em}
\end{figure*}

\section{Proposed Method}

\subsection{Overview}

The definition of image harmonization is to input a composite image with the corresponding foreground mask and output the harmonious image. Formulaically, the composite image $I_c$ consists of the foreground image $I_f$ and the background image $I_b$. The foreground mask $M_f$ denotes the foreground region in the composite image $I_c$. The background mask $M_b = 1 - M_f$ accordingly. The goal of image harmonization network is training a generator $G$ to generate the harmonious image $\hat{I}$, where $\hat{I} = G(I_c, M_f)$. If there is a ground-truth image $I$ of the composite image $I_c$, the optimization direction of the model $G$ is to make $\hat{I}$ close to $I$.

Our proposed fine-grained region-aware image harmonization framework is illustrated in Figure \ref{fig:fig2}, which is a two-stage network. We use a simple U-Net \cite{ronneberger2015u} alike network as the base network in the first stage. In this stage, we feed the composite image and the foreground mask into the encoder-decoder network to obtain the global coarsely adjusted image. In the second stage, we first cluster the foreground mask into several submasks adaptively. Then, each submask and the global coarsely adjusted image generated in the first stage are concatenated respectively and fed into the lightweight cascaded module, which makes full use of the region-aware local feature according to the submasks. Finally, we embeds the fusion prediction module into the decoder of the lightweight cascaded module, to ultilize the different levels of the harmonization results comprehensively and generate the final refined harmonious images. The whole FRIH network is trained end-to-end.

\subsection{Base Network}

The base network inputs the composite image and the foreground mask, and outputs the global coarsely adjusted image. The base network is a simple U-Net \cite{ronneberger2015u} alike network, including an encoder and a decoder. The encoder has been downsampled for 7 times and the decoder has 7 deconvolution layers correspondingly. There is a skip connection between each convolution layer in the encoder and the corresponding deconvolution layer with the same feature map size in the decoder.   

\subsection{Submask Extraction}

A significant step of FRIH is to extract the submasks of the global mask. The number $K$ of the submasks needs to be adaptive for different input images. Therefore, We apply CFSFDP clustering algorithm \cite{rodriguez2014clustering} to extract $K$ submasks for each global mask. The original CFSFDP algorithm has two basic ideas, which are that cluster centers have a higher density than their neighbors, and are at a relatively large distance from points with higher densities \cite{rodriguez2014clustering}. In our method, we cluster the pixels by their corresponding pixel RGB values in the composite image. The local density $\rho _i$ of pixel $p_i$ is measured as:
\begin{equation}
\begin{aligned}
\rho _i=\sum_{j}{\chi (d_{ij}-d_c)}
\label{eq:density}
\end{aligned}
\end{equation}
where $d_{ij}$ is the normalized euclidean distance between the $(r,g,b)$ vectors of pixels $p_i$ and $p_j$. $d_{c}$ is the cutoff distance. We will introduce the setting of $d_c$ in Section \ref{sec:experiment}. $\chi (x) = 1$ if $x < 0$ and $\chi (x) = 0$ otherwise. The minimum distance $\delta _i$ between pixel $p_i$ and any other pixel $p_j$ with higher density is defined as:
\begin{equation}
\begin{aligned}
\delta _i=\min\limits_{j:\rho _j>\rho _i}(d_{ij})
\label{eq:similarity}
\end{aligned}
\end{equation}

Note that if $p_i$ has the highest density in the image, then $\delta _i=\max_j(d_{ij})$. In our method, according to the statistical observations on a large number of images, we find that the appropriate number of submasks for the foreground will not exceed 10 in the vast majority of cases. Therefore, we sort all pixels by $\delta$. The 10 RGB values with the highest $\delta$ are considered to be the candidate cluster centers. Note that the  RGB values of different candidate cluster centers should be different. For example, if all the pixels in the foreground have the same RGB value, there will be only 1 candidate cluster centers. The candidate cluster centers whose densities $\rho$ do not exceed 10 will be considered as isolated outliers. Only the candidate cluster centers whose 
densities are higher than 10 are considered as the final cluster centers. Thus the cluster centers of each foreground mask are obtained adaptively. The number of the cluster centers $K$ is in the range of 1 to 10.

\begin{figure}[!t]
\centering
\includegraphics[width=\columnwidth, height=0.6\columnwidth]{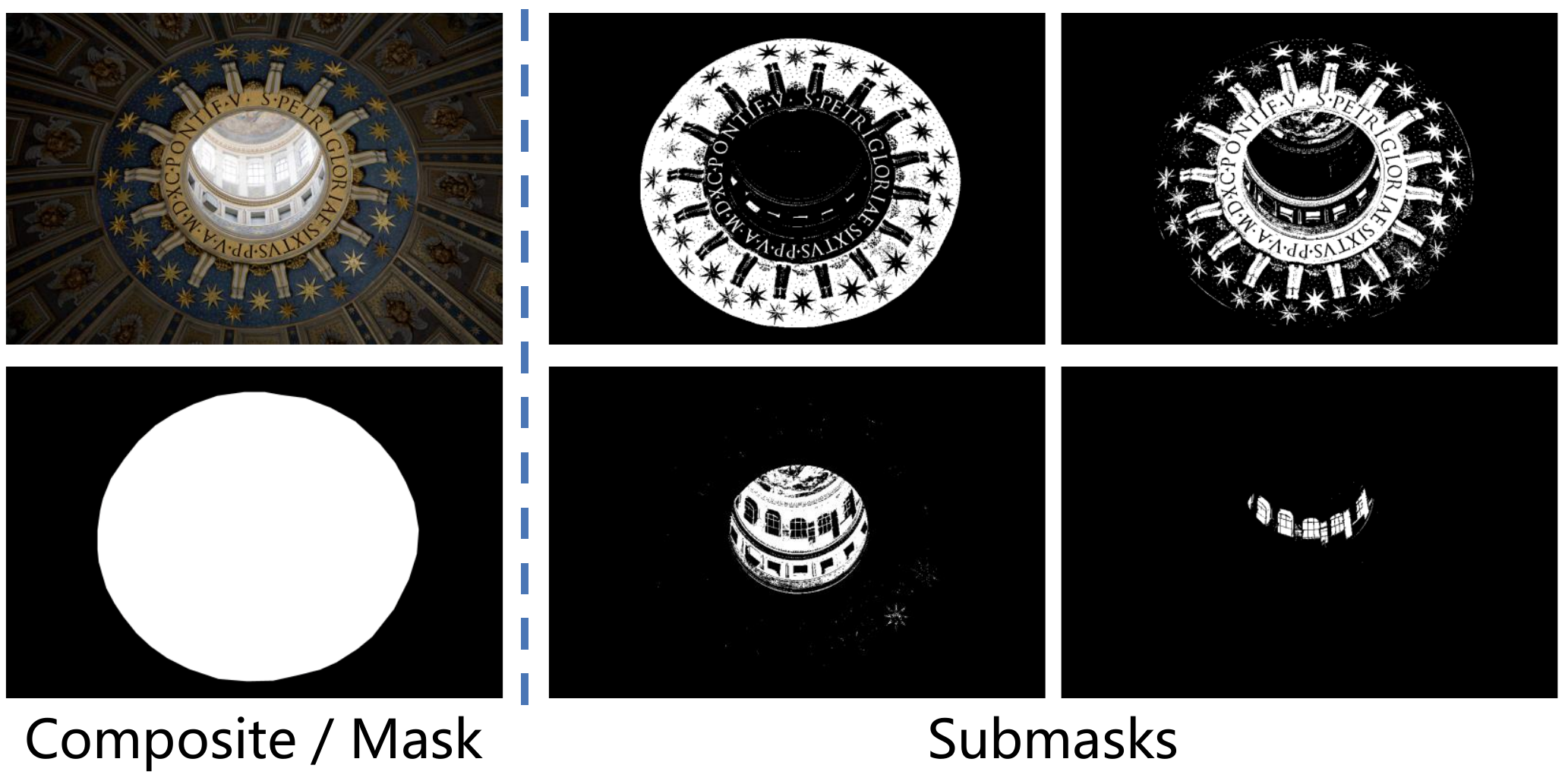}
\caption{A representative case of the submask extraction. The origin mask (left) is divided into 4 submasks (right).}
\label{fig:fig3}
\end{figure}


When all the cluster centers are obtained, the remaining pixels are assigned to the same cluster as their nearest pixel (measured by the normalized euclidean distance of RGB values) of higher density.

In this way, all the pixels in the foreground mask $M_f$ are assigned into $K$ clusters. For each composite image, we obtain $K$ submasks $Subm^1$,$Subm^2$,...,$Subm^k$. Figure \ref{fig:fig3} displays a representative set of the generated submasks.  In this case, $K=4$.


\subsection{Lightweight Cascaded Module}

In this module, we concatenate the global coarsely adjusted image generated in the first stage and each submask respectively. All the concatenated pairs are fed into the lightweight cascaded module. We imitate the structure of U-Net \cite{ronneberger2015u} to construct the lightweight cascaded module. In the cascaded encoder, seven $4\times4$ convolution layers are used to extract different level features of the coarsely adjusted image. In the cascaded decoder part, different from the original decoder in U-Net, at each layer, we use a $1\times1$ convolution layer to fuse the features from three sources: the previous decoder layer,  the corresponding encoder layer in the first stage (the yellow arrow in Figure \ref{fig:fig2}), the corresponding encoder layer in the cascaded module. In this way, we not only further harmonize the image based on the results from first stage network, but also utilize the features from the original composite image. The details of this module are shown in Figure \ref{fig:fig2}. Thus the output of the $i$-th cascaded decoder layer $DC_i$ can be calculated by the following equation:

\begin{equation}
    ET_i=Conv_{trans}(E_i)
\end{equation}

\begin{equation}
    F_i=Conv_{fuse}([DC_{i-1},ET_i,EC_i])
\end{equation}

\begin{equation}
    DC_i=Conv_{up}(F_i)
\end{equation}
where $EC_i$ denotes the output of the $i$-th layer in the cascaded encoder, which contains the information from the coarsely adjusted image. $DC_i$ denotes the output of the $i$-th layer in the cascaded decoder. $E_i$ denotes the features from the first stage encoder. We use two $3\times3$ convolution layers as transfer function $Conv_{trans}$ to transfer this feature $E_i$ into $ET_i$ to make it suitable for the cascaded module. The operator [] means the concatenation operation and $Conv_{fuse}$ is a $1\times1$ convolution layer to fuse the features from three different sources. $Conv_{up}$ is a $4\times4$ transpose convolution layer to upsample and decode the fused feature $F_i$. In this way, we obtain the cascaded adjust features from the last layer of our cascaded module. Note that the channels of the features are much smaller than those in the original U-Net. The cascaded module together with the following embedded fusion prediction module has only 2.68 M parameters, only accounting for 22.4\% of the parameters of the whole FRIH network. Thus, our cascaded module is lightweight.

\begin{figure}[!t]
\centering
\includegraphics[width=\columnwidth]{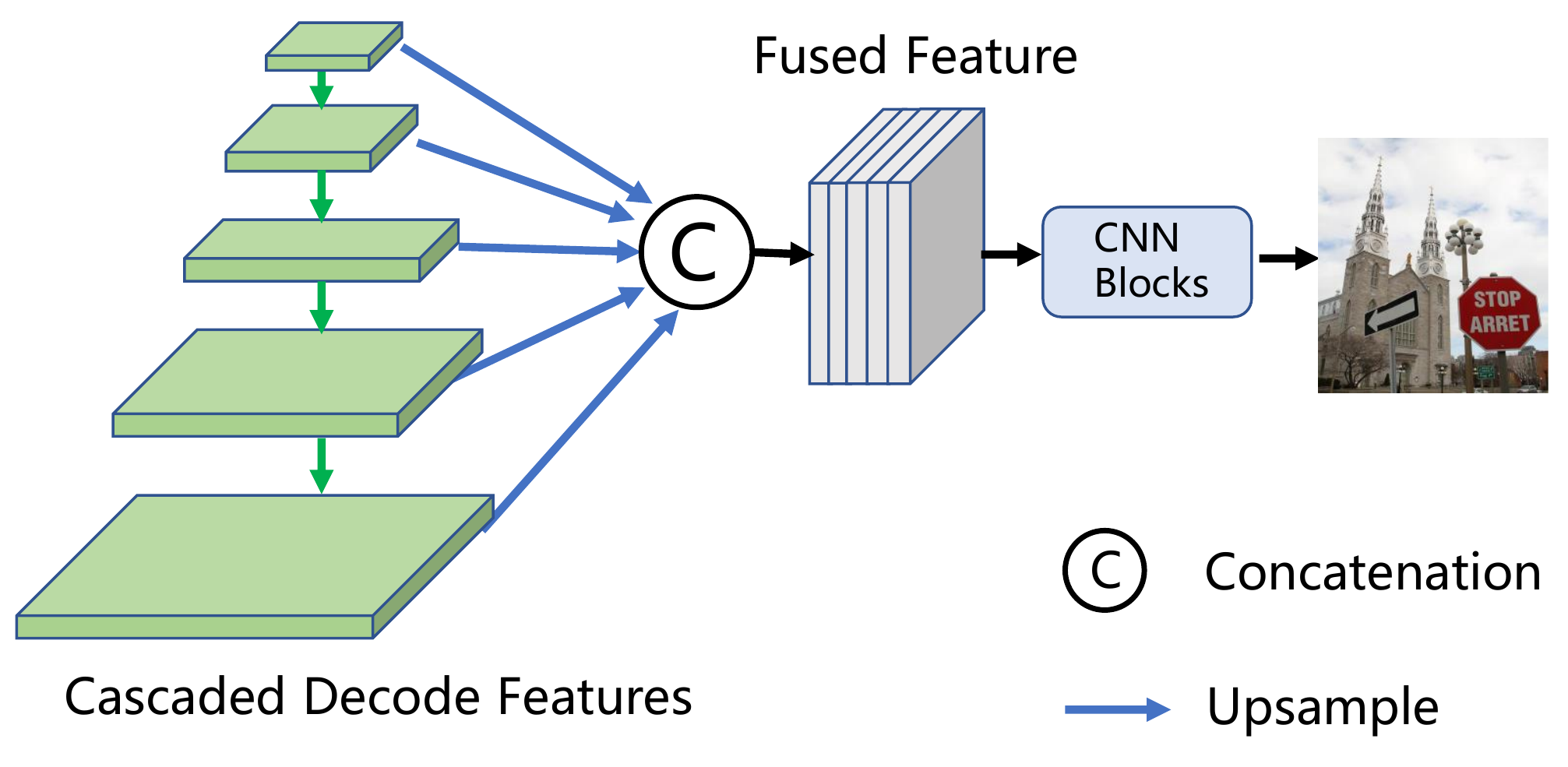}
\caption{The fusion prediction module. Features from all the cascaded decoder layers are used to predict results.}
\label{fig:fig_fusion}
\end{figure}

\subsection{Fusion Prediction}

\begin{table*}[t]
\caption{Comparison of image harmonization results in each sub-dataset on iHarmony4 test set.}
\normalsize
\centering
\begin{tabular}{c|cc|cc|cc|cc|cc}
\toprule[1.5pt]
Sub-dataset               & \multicolumn{2}{c|}{HCOCO} & \multicolumn{2}{c|}{HAdobe5k} & \multicolumn{2}{c|}{HFlickr} & \multicolumn{2}{c|}{Hday2night} & \multicolumn{2}{c}{All} \\
Metric & MSE$\downarrow$ & PSNR$\uparrow$ & MSE$\downarrow$ & PSNR$\uparrow$ & MSE$\downarrow$ & PSNR$\uparrow$ & MSE$\downarrow$ & PSNR$\uparrow$ & MSE$\downarrow$ & PSNR$\uparrow$ \\
\midrule[0.8pt]
Input composite          & 69.37 & 33.94    & 345.54 & 28.16 & 264.35 & 28.32 & 109.65 & 34.01 & 172.47 & 31.63      \\
Lalonde \cite{lalonde2007using}            & 110.10 & 31.14    & 158.90 & 29.66 & 329.87 & 26.43 & 199.93 & 29.80 & 150.53 & 30.16      \\
Xue \cite{xue2012understanding}      & 77.04 & 33.32    & 274.15 & 28.79 & 249.54 & 28.32 & 190.51 & 31.24 & 155.87 & 31.40      \\
Zhu \cite{zhu2015learning}      & 79.82 & 33.04    & 414.31 & 27.26 & 315.42 & 27.52 & 136.71 & 32.32 & 204.77 & 30.72      \\
DIH \cite{tsai2017deep}       & 51.85 & 34.69    & 92.65 & 32.28 & 163.38 & 29.55 & 82.34 & 34.62 & 76.77 & 33.41      \\
S$^2$AM \cite{cun2020improving}        & 41.07 & 35.47    & 63.40 & 33.77 & 143.45 & 30.03 & 76.61 & 34.50 & 59.67 & 34.35     \\
DoveNet \cite{cong2020dovenet} & 36.72 & 35.83    & 52.32 & 34.34 & 133.14 & 30.21 & 54.05 & 35.18 & 52.36 & 34.75      \\
ADFM \cite{hao2020image} & - & 36.87    & - & 34.99 & - & 33.36 & - & 34.31 & - & 35.86      \\
BargainNet \cite{cong2021bargainnet} & 24.84 & 37.03    & 39.94 & 35.34 & 97.32 & 31.34 & 50.98 & 35.67 & 37.82 & 35.88      \\
IIH \cite{guo2021intrinsic} & 24.92 & 37.16    & 43.02 & 35.20 & 105.13 & 31.34 & 55.53 & 35.96 & 38.71 & 35.90      \\
RainNet \cite{ling2021region} & - & 37.08    & - & 36.22 & - & 31.64 & - & 34.83 & - & 36.12      \\
iDIH \cite{sofiiuk2021foreground} & 19.29 & 38.44    & 30.87 & 36.09 & 84.10 & 32.61 & 55.24 & 37.26 & 30.56 & 37.08      \\
D-HT \cite{guo2021image} & 16.89 & 38.76    & 38.53 & 36.88 & 74.51 & 33.13 & 53.01 & 37.10 & 30.30 & 37.55      \\
\midrule[0.8pt]
FRIH (ours)           & \textbf{15.05} &  \textbf{39.35}  & \textbf{23.61} & \textbf{37.69} & \textbf{68.93} & \textbf{33.48} & \textbf{42.78} & \textbf{37.89} & \textbf{23.98}  & \textbf{38.19}        \\
\bottomrule[1.5pt]
\end{tabular}
\label{tab:sota}
\vspace{.5cm}
\end{table*}

\begin{table*}[t]
\caption{Comparison of image harmonization results in each foreground ratio range on iHarmony4 test set.}
\normalsize
\centering
\begin{tabular}{c|cc|cc|cc|cc}
\toprule[1.5pt]
Foreground ratios               & \multicolumn{2}{c|}{0\%-5\%} & \multicolumn{2}{c|}{5\%-15\%} & \multicolumn{2}{c|}{15\%-100\%} & \multicolumn{2}{c}{All} \\
Metric & MSE$\downarrow$ & fMSE$\downarrow$ & MSE$\downarrow$ & fMSE$\downarrow$ & MSE$\downarrow$ & fMSE$\downarrow$ & MSE$\downarrow$ & fMSE$\downarrow$ \\
\midrule[0.8pt]
Input composite          & 28.51 & 1208.86    & 119.19 & 1323.23 & 577.58 & 1887.05 & 172.47 & 1387.30      \\
Lalonde \cite{lalonde2007using}            & 41.52 & 1481.59    & 120.62 & 1309.79 & 444.65 & 1467.98 & 150.53 & 1433.21      \\
Xue \cite{xue2012understanding}      & 31.24 & 1325.96    & 132.12 & 1459.28 & 479.53 & 1555.69 & 155.87 & 1411.40      \\
Zhu \cite{zhu2015learning}      & 33.30 & 1297.65    & 145.14 & 1577.70 & 682.69 & 2251.76 & 204.77 & 1580.17      \\
DIH \cite{tsai2017deep}       & 18.92 & 799.17    & 64.23 & 752.86 & 228.86 & 768.89 & 76.77 & 773.18      \\
S$^2$AM \cite{cun2020improving}        & 15.09 & 623.11    & 48.33 & 540.54 & 177.62 & 592.83 & 59.67 & 594.67    \\
DoveNet \cite{cong2020dovenet} & 14.03 & 591.88    & 44.90 & 504.42 & 152.07 & 505.82 & 52.36 & 549.96      \\
BargainNet \cite{cong2021bargainnet} & 10.55 & 450.33    & 32.13 & 359.49 & 109.23 & 353.84 & 37.82 & 405.23     \\
RainNet \cite{ling2021region} & 11.66 & 550.38    & 32.05 & 378.69 & 117.41 & 389.80 & 40.29 & 469.60   \\
iDIH \cite{sofiiuk2021foreground} & 8.38 & 366.32    & 25.39 & 287.02 & 89.44 & 297.94 & 30.56 & 330.45      \\
\midrule[0.8pt]
FRIH (ours)           & \textbf{6.89} &  \textbf{305.28} & \textbf{19.88} & \textbf{226.45} & \textbf{70.05} & \textbf{205.83} & \textbf{23.98} & \textbf{252.63}        \\
\bottomrule[1.5pt]
\end{tabular}
\label{tab:ratio}
\vspace{.5cm}
\end{table*}

In order to utilize the different levels of harmonization results comprehensively, we further designed the fusion prediction module. In the previous image harmonization methods and traditional U-Net, they only use the features from the last decoder layer to predict the generated images. This is because that the features from the former decoder layers have not been adjust to be close to the target enough. However, in our cascaded module, the input is the coarsely adjusted image. Therefore, all the cascaded decoder layers have already fused the information of the coarsely harmonized image. The outputs from different cascaded decoder layers represent that the features of images that are adjusted to varying degrees. The similarity between the composite image and the real ground-truth image varies among different sub-regions. Some sub-regions in the composite image are relatively similar to the real ground-truth image, while others may be much far away from the target. Thus different sub-regions need different degrees of harmonization. As shown in Figure \ref{fig:fig_fusion}, we fuse features from all the cascaded decoder layers together to predict the harmonized result, which enables the prediction head to utilize features from different harmonization levels. Since the resolution of the features from different cascaded decoder layers are different, we upsample all these feature maps to the resolution $256\times256$ and concatenate them together. Then we use two $3\times3$ and one $1\times1$ convolution layers to convert these fused feature maps to a 3-channel RGB image.

As shown in Figure \ref{fig:fig2}, we can obtain $K$ refined harmonious images from the input K pairs. The final output is generated by combining these images according to the corresponding submasks.

\subsection{Training Loss}





Since we use base network to predict coarse results and a cascaded module to obtain refined images, for each image, the loss function $L_{total}$ of FRIH consists of two parts, which are $L_{coarse}$ and $L_{refine}$. we use the following equation to calculate $L_{total}$.
\begin{equation}
    L_{total}=L_{coarse}+L_{refine}
\end{equation}
\begin{equation}
    L_{coarse}=\sum_{h,w}\frac{||I_{h,w}-\hat{I}_{h,w}||^2_2}{max(Area_{mask},A_{min})}
\label{eq:coarse}
\end{equation}
\begin{equation}
    L_{refine}=\sum_{i=1}^{K}\sum_{h,w}\frac{||I_{h,w}-\hat{I}_{h,w}||^2_2 \cdot Subm_{h,w}^{i}}{max(Area_{Subm^{i}},A_{min})}
\label{eq:refine}
\end{equation}

where $\hat{I}$ is the prediction result. In equations \ref{eq:coarse} and \ref{eq:refine}, $\hat{I}$ represents the coarsely adjusted image in the first stage and the final output image in the second stage, respectively. It should be noted that in $L_{refine}$, we only focus on the submask area, so the loss is multiplied by the submask. Furthermore, We find that the images with small foreground masks are always hard examples and we want our model to learn more information from them. Therefore, we divide the loss by the area of the foreground mask. $A_{min}$ is a constant, which is set to 100 in all the experiments. For those foreground masks whose areas are smaller than $A_{min}$, we treat them as $A_{min}$.

\section{Experiments}\label{sec:experiment}

\begin{table*}[t]
\caption{Ablation study in each sub-dataset on iHarmony4 test set.}
\normalsize
\centering
\begin{tabular}{c|cc|cc|cc|cc|cc}
\toprule[1.5pt]
Sub-dataset               & \multicolumn{2}{c|}{HCOCO} & \multicolumn{2}{c|}{HAdobe5k} & \multicolumn{2}{c|}{HFlickr} & \multicolumn{2}{c|}{Hday2night} & \multicolumn{2}{c}{All} \\
Metric & MSE$\downarrow$ & PSNR$\uparrow$ & MSE$\downarrow$ & PSNR$\uparrow$ & MSE$\downarrow$ & PSNR$\uparrow$ & MSE$\downarrow$ & PSNR$\uparrow$ & MSE$\downarrow$ & PSNR$\uparrow$ \\
\midrule[0.8pt]
Input composite          & 69.37 & 33.94    & 345.54 & 28.16 & 264.35 & 28.32 & 109.65 & 34.01 & 172.47 & 31.63      \\
Baseline           & 21.38 &  38.24        & 37.15 & 35.79 & 90.12 & 32.25 & 50.12 & 37.45 & 35.38  & 36.82        \\
Baseline+Cascade           & 16.52 &  39.09        & 26.21 & 37.42 & 75.82 & 33.05 & \textbf44.21 & 37.78 & 26.31  & 37.90        \\
Baseline+Cascade+Fusion           & \textbf{15.05} &  \textbf{39.35}  & \textbf{23.61} & \textbf{37.69} & \textbf{68.93} & \textbf{33.48} & \textbf{42.78} & \textbf{37.89} & \textbf{23.98}  & \textbf{38.19}       \\
\bottomrule[1.5pt]
\end{tabular}
\label{tab:ablation}
\vspace{.5cm}
\end{table*}

\begin{table*}[t]
\caption{Ablation study in each foreground ratio range on iHarmony4 test set.}
\normalsize
\centering
\begin{tabular}{c|cc|cc|cc|cc}
\toprule[1.5pt]
Foreground ratios               & \multicolumn{2}{c|}{0\%-5\%} & \multicolumn{2}{c|}{5\%-15\%} & \multicolumn{2}{c|}{15\%-100\%} & \multicolumn{2}{c}{All} \\
Metric & MSE$\downarrow$ & fMSE$\downarrow$ & MSE$\downarrow$ & fMSE$\downarrow$ & MSE$\downarrow$ & fMSE$\downarrow$ & MSE$\downarrow$ & fMSE$\downarrow$ \\
\midrule[0.8pt]
Input composite          & 28.51 & 1208.86    & 119.19 & 1323.23 & 577.58 & 1887.05 & 172.47 & 1387.30      \\
Baseline           & 10.21 &  421.53        & 29.68 & 336.32 & 109.21 & 345.19 & 35.38 & 380.12        \\
Baseline+Cascade           & 7.28 &  322.86        & 21.42 & 244.15 & 77.82 & 230.47 & 26.31 & 268.52       \\
Baseline+Cascade+Fusion           & \textbf{6.89} &  \textbf{305.28} & \textbf{19.88} & \textbf{226.45} & \textbf{70.05} & \textbf{205.83} & \textbf{23.98} & \textbf{252.63} \\
\bottomrule[1.5pt]
\end{tabular}
\label{tab:ablation_ratio}
\vspace{.5cm}
\end{table*}

\subsection{Datasets and Evaluation Metrics}

To demonstrate the effectiveness of our FRIH, we conduct experiments on the public image harmonization dataset iHarmony4 \cite{cong2020dovenet}. This dataset consists of four sub-datasets, including HCOCO, HAdobe5k, HFlickr and Hday2night. There are totally 65,742 training image pairs and 7,407 test image pairs in iHarmony4. All the image pairs are generated by modifying the specific foreground regions of the normal images, which are converted to corresponding the inharmonious images in this way. We follow the same train-test split as DoveNet \cite{cong2020dovenet} in the experiments.

Following prior work \cite{tsai2017deep,cong2020dovenet}, we use Mean Squared Error (MSE) scores and Peak Signal-to-Noise Ratio (PSNR) scores on RGB channels to evaluate the image harmonization performance. Besides, to eliminate the influence of the foreground area size on the metric, we also introduced the foreground Mean Squared Error (fMSE) score \cite{cong2020dovenet}. Note that all the input and output images are resized to 256 $\times$ 256 resolution during training and test, thus all the evaluation metrics are calculated based on the 256 $\times$ 256 resolution.

\subsection{Implementation details}
We use Adam Optimizer with $\beta_1 = 0.9$ and $\beta_2 = 0.999$ to train our model for 180 epochs on 8 Tesla V100 GPUs. The initial learning rate is 0.008, which decays by 10 at epoch 160 and 175. The batchsize is 128. All the images are resized to $256\times256$ in both training and test process. We use horizontal flip and random size crop to augment the data during training. The whole FRIH network is trained end-to-end. The cutoff distance $d_c$ is set to 0.1.

\subsection{Comparison with the State-of-the-Art}

We compare our FRIH with other image harmonization methods on iHarmony4. Table \ref{tab:sota} and Table \ref{tab:ratio} separately show the results in each sub-dataset and each foreground ratio range. From Table \ref{tab:sota} and Table \ref{tab:ratio} we can find that:

(1) Our method achieves 38.19 PSNR and 23.98 MSE on iHarmony4 test set, which performs better than all the other image harmonization methods on iHarmony4 test set in a large margin. Compared to the previous SOTA methods, the PSNR of our method has an obvious improvement. The latest transformer-based method D-HT \cite{guo2021image} achieved 37.55 using a large model. While our method is the only one whose PSNR is larger than 38 dB, which proves that the fine-grained region-aware framework works well. It should be noted that iDIH \cite{sofiiuk2021foreground} obtains higher performance than 37.08 dB when adding extra pre-trained semantic segmentation model, which all other methods do not use. It is unfair to make comparison so that we only show their results without the extra model.   

(2) In Table \ref{tab:sota}, our method performs much better than existing methods on all of the 4 sub-datasets, demonstrating the robustness of our cascaded module and fusion prediction strategy.

(3) In Table \ref{tab:ratio}, our method performs better than all the existing methods on the composite images with different foreground ratio, proving that our submask extraction is well adaptive regardless of the foreground area size.

\subsection{Ablation Study}

We compare the following models on iHarmony4 dataset to show the effectiveness of FRIH's parts:
\begin{itemize}
  \item \emph{Input composite}. It means that we directly use the input composite image as the final output result, without any harmonization processing.
  \item \emph{Baseline}. The baseline only uses the base network to make a coarse-grained harmonization, without the submask extraction and the lightweight cascaded module.
  \item \emph{Baseline+Cascade}. We extract submasks adaptively for each foreground mask and feed the concatenation of each submask and the coarsely adjusted image into the lightweight cascaded module.
  \item \emph{Baseline+Cascade+Fusion (FRIH)}. This is the full version of our algorithm, including the submask extraction, lightweight cascaded module together with the fusion prediction strategy. We use features from all the cascaded decoder layers to predict the final harmonized results.
\end{itemize}


\begin{table}
\caption{Comparative experiment in terms of $d_c$ on iHarmony4 test set.}
\normalsize
\centering
\begin{tabular}{c|cccccc}
\toprule[1.5pt]
$d_c$ & 0.01 & 0.05 & 0.1 & 0.2 & 0.3 & 0.4\\
\midrule[0.8pt]
MSE$\downarrow$ & 25.26 & 24.49 & \textbf{23.98} & 24.76 & 25.41 & 25.90\\
PSNR$\uparrow$ & 38.09 & 38.17 & \textbf{38.19} & 38.15 & 38.08 & 38.02\\
\bottomrule[1.5pt]
\end{tabular}
\label{tab:cluster}
\vspace{.1cm}
\end{table}

The ablation study results in each sub-dataset and each foreground ratio range on iHarmony4 are presented in Table \ref{tab:ablation} and Table \ref{tab:ablation_ratio} respectively, showing that:

(1) Both the lightweight cascaded module and fusion prediction strategy can increase PSNR and decrease MSE. \emph{Baseline+Cascade} performs significantly better \emph{Baseline} with the gain of 1.08 dB in PSNR. \emph{Baseline+Cascade+Fusion} outperforms \emph{Baseline+Cascade} with the gain of 0.29 dB in PSNR. It demonstrates the effectiveness of both the lightweight cascaded module and the fusion prediction strategy. Compared to \emph{Baseline}, our FRIH has a gain of 1.37 dB in PSNR and a decrease of 11.40 in MSE metric, proving that the whole global-local framework is effective and all modules are reciprocal.

(2) The effectiveness of the lightweight cascaded module is not very apparent on Hday2night sub-dataset. We analyze the composite images and the foreground masks. We find that the foreground areas in Hday2night are always sky, water surface or other similar categories with single and pure color. These areas are homogeneous. Therefore, it is hard to separate these foreground areas into different sub-regions, making our cascaded module useless. From another perspective, it further proves the effectiveness of our method on foregrounds with different appearance patterns. By the way, our fusion prediction strategy still works on Hday2night, since it is not affected by submasks.

(3) The increase of PSNR between \emph{Baseline} and \emph{Baseline+Cascade} on HAobe5k sub-dataset achieves 1.63, which is the largest among 4 sub-datasets in Table \ref{tab:ablation}. We think the reason is that the foreground masks in HAdobe5k are much bigger than other three sub-datasets, which means these bigger foreground masks are more likely to contain disparate sub-regions or different appearance patterns. The baseline model treats them in the same way while the cascaded module can adjust them adaptively, leading to a significantly increase of PSNR and decrease of MSE.

(4) The increase between \emph{Baseline} and \emph{Baseline+Cascade} of images whose foreground mask sizes are larger than 15\% is also the biggest in Table \ref{tab:ablation_ratio}. We believe the reason is as same as the reason why the increase in the sub-dataset HAdobe5k is the biggest. Both (3) and (4) can prove that our submasks-based cascaded module is efficient for image harmonization task, especially for those images with large foreground masks.

Moreover, the cutoff distance $d_c$ in the submask extraction module is a significant hyper-parameter in our method. It determines the granularity of the submasks division. We conduct comparative experiment in terms of $d_c$ on iHarmony4 test set. The results are shown in Table \ref{tab:cluster}. $d_c$ is set to 0.01, 0.05, 0.1, 0.2, 0.3 and 0.4, respectively. When $d_c=0.1$, our FRIH achieves the best performance. Thus we set $d_c=0.1$ in all the other experiments in this paper. When $d_c=0.01$, the performance is lower than $d_c=0.1$, because the clustering is so fine-grained that many noisy isolated outliers are generated and have negative effect on the selection of cluster centers. When $d_c=0.3$ and $d_c=0.4$, the performance also drops compared with $d_c=0.1$, which is caused by the too coarse-grained clustering. Several color blocks, which are not similar, are clustered into a same cluster. However, The PSNR in all these experiments exceeds 38, demonstrating the robustness of our method.


\begin{table}
\caption{Comparison of Model Size and GFLOPS.}
\normalsize
\centering
\begin{tabular}{c|ccc}
\toprule[1.5pt]
Method & Params$\downarrow$ & GFLOPS$\downarrow$ & PSNR$\uparrow$\\
\midrule[0.8pt]
DoveNet \cite{cong2020dovenet} & 54.76 & 18.97 & 34.75\\
IIH \cite{guo2021intrinsic} & 40.83 & 196.3 &35.90\\
RainNet \cite{ling2021region} & 54.75 & \textbf{18.95} & 36.12\\
\midrule[0.8pt]
FRIH (ours) & \textbf{11.98} &  25.12 & \textbf{38.19}\\
\bottomrule[1.5pt]
\end{tabular}
\label{tab:time}
\vspace{.5cm}
\end{table}

\subsection{Comparison of Model Size and Computation}
We also compare the model size and computation of our method FRIH with existing SOTA methods. As shown in Table \ref{tab:time}, compared to existing methods, our FRIH method achieves much higher results while the model size is only about 1/5 of theirs. The computation cost of our model is also close to the computation of DoveNet \cite{cong2020dovenet} and RainNet \cite{ling2021region}, and much lower than that of IIH \cite{guo2021intrinsic}.

\begin{table}[t]
\caption{Comparison between our FRIH and other state-of-the-art methods under user study.}
\normalsize
\centering
\begin{tabular}{c|cccc}
\toprule[1.5pt]
Method & Input & DoveNet & RainNet & FRIH (Ours) \\
\midrule[0.8pt]
Votes$\uparrow$ & 261 & 485 & 501 & \textbf{733}\\
Preference$\uparrow$ & 12.1\% & 23.2\% & 24.1\% & \textbf{40.6\%} \\
\bottomrule[1.5pt]
\end{tabular}
\label{tab:userstudy}
\vspace{.5cm}
\end{table}

\subsection{User Study}

\begin{figure*}[t]
\centering
\includegraphics[width=0.94\textwidth]{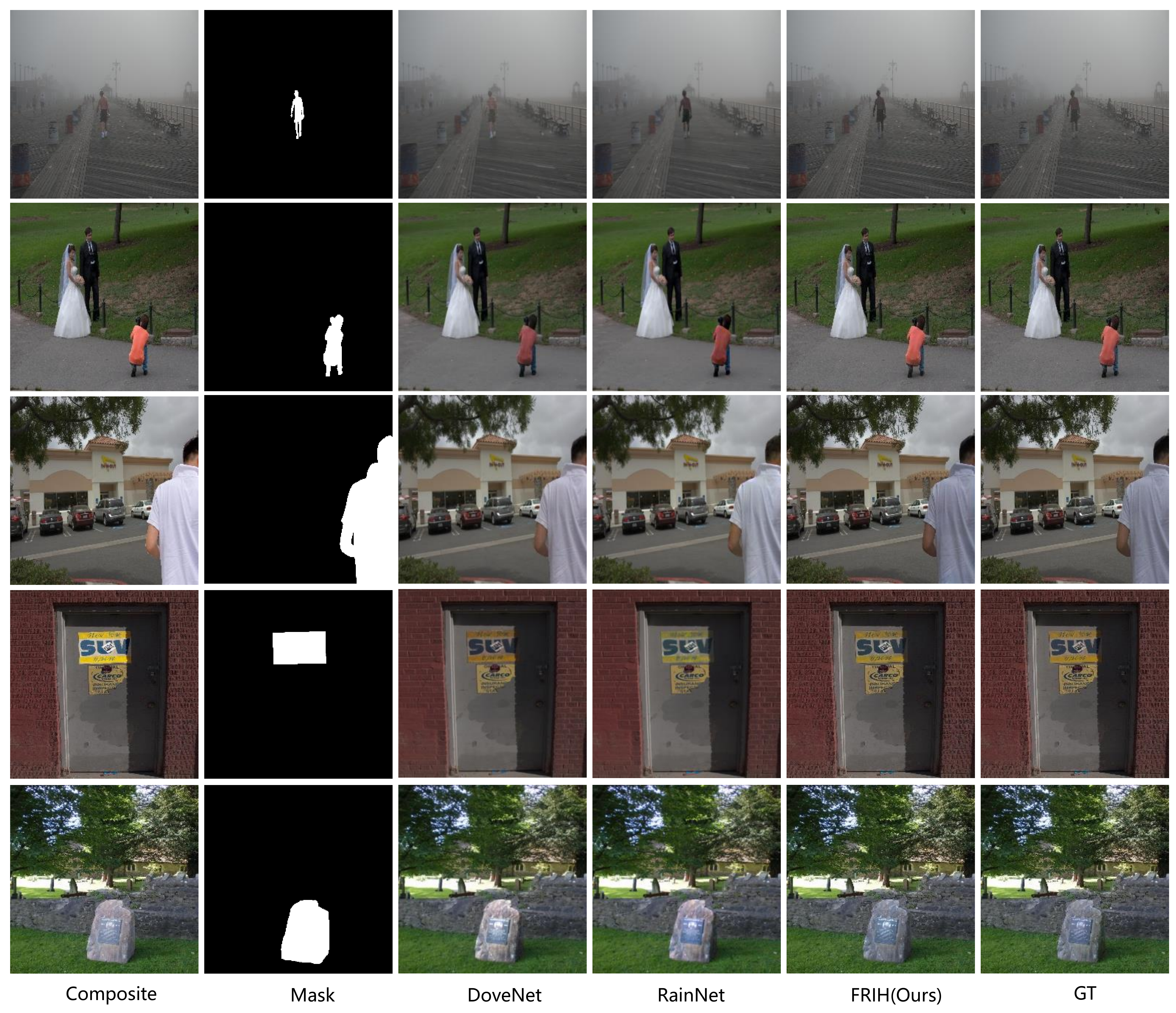}
\caption{Qualitative comparison between our FRIH and other state-of-the-art methods.}
\label{fig:visual}
\end{figure*}

Sometimes the PSNR and MSE metrics can not correctly reflect the impressions of human eyes. Thus, we conduct the user study on 99 images dataset provided by DIH \cite{tsai2017deep}. We invite 20 volunteers and ask them to select the most realistic images they think from composite images and results generated by DoveNet \cite{cong2020dovenet}, RainNet \cite{ling2021region} and our FRIH. The total number of votes are 1980. As shown in Table \ref{tab:userstudy}, our FRIH attains much more votes than other state-of-the-art methods, which proves that our method not only performs best on PSNR and MSE metrics, but also can generate more realistic images than other methods according to human judgement.

\subsection{Qualitative Analysis}

We show some visualization results on iHarmony4 test dataset of our FRIH method against DoveNet \cite{cong2020dovenet} and RainNet \cite{ling2021region} in Figure \ref{fig:visual}. In all the 5 cases, the harmonization images of our FRIH are more closer to the ground-truth images compared to DoveNet \cite{cong2020dovenet} and RainNet \cite{ling2021region}. The foreground in each case has different appearance patterns. Applying DoveNet and RainNet, only part of the foreground has satisfactory harmonization performance. For example, in the fourth line, the blue word 'SUV' is handled well while the yellow poster is not, both in DoveNet and RainNet. However, our FRIH algorithm handles the details of all the sub-regions better, demonstrating the effectiveness of our fine-grained region-aware framework. Some representative visualization results of out FRIH method against DoveNet \cite{cong2020dovenet} and RainNet \cite{ling2021region} on the 99 image test dataset \cite{tsai2017deep} are presented in the supplementary materials.


\section{Conclusion}

We proposed a two-stage fine-grained region-aware image harmonization framework, which is simple, novel and effective. In the first stage, the whole input foreground mask is used to make a global coarse-grained harmonization. In the second stage, we adaptively cluster the input foreground mask into several submasks by the corresponding pixel RBG values. Each submask and the coarsely adjusted image are concatenated respectively and fed into the lightweight cascaded module. Moreover, we further designed a fusion prediction module by fusing features from all the cascaded decoder layers together to generate the final result. It addresses the problem that existing methods ignores the difference of each color block and loses some specific details. Extensive experimental results demonstrate the effectiveness and efficiency of our FRIH algorithm and its superiority over the state-of-the-art competitors on iHarmony4.

\bibliographystyle{ACM-Reference-Format}
\bibliography{sample-base}


\begin{thebibliography}{40}


\ifx \showCODEN    \undefined \def \showCODEN     #1{\unskip}     \fi
\ifx \showDOI      \undefined \def \showDOI       #1{#1}\fi
\ifx \showISBNx    \undefined \def \showISBNx     #1{\unskip}     \fi
\ifx \showISBNxiii \undefined \def \showISBNxiii  #1{\unskip}     \fi
\ifx \showISSN     \undefined \def \showISSN      #1{\unskip}     \fi
\ifx \showLCCN     \undefined \def \showLCCN      #1{\unskip}     \fi
\ifx \shownote     \undefined \def \shownote      #1{#1}          \fi
\ifx \showarticletitle \undefined \def \showarticletitle #1{#1}   \fi
\ifx \showURL      \undefined \def \showURL       {\relax}        \fi
\providecommand\bibfield[2]{#2}
\providecommand\bibinfo[2]{#2}
\providecommand\natexlab[1]{#1}
\providecommand\showeprint[2][]{arXiv:#2}

\bibitem[Azadi et~al\mbox{.}(2020)]%
        {azadi2020compositional}
\bibfield{author}{\bibinfo{person}{Samaneh Azadi}, \bibinfo{person}{Deepak
  Pathak}, \bibinfo{person}{Sayna Ebrahimi}, {and} \bibinfo{person}{Trevor
  Darrell}.} \bibinfo{year}{2020}\natexlab{}.
\newblock \showarticletitle{Compositional gan: Learning image-conditional
  binary composition}.
\newblock \bibinfo{journal}{\emph{International Journal of Computer Vision}}
  \bibinfo{volume}{128}, \bibinfo{number}{10} (\bibinfo{year}{2020}),
  \bibinfo{pages}{2570--2585}.
\newblock


\bibitem[Cai and Vasconcelos(2018)]%
        {cai2018cascade}
\bibfield{author}{\bibinfo{person}{Zhaowei Cai} {and} \bibinfo{person}{Nuno
  Vasconcelos}.} \bibinfo{year}{2018}\natexlab{}.
\newblock \showarticletitle{Cascade r-cnn: Delving into high quality object
  detection}. In \bibinfo{booktitle}{\emph{Proceedings of the IEEE conference
  on computer vision and pattern recognition}}. \bibinfo{pages}{6154--6162}.
\newblock


\bibitem[Chen and Kae(2019)]%
        {chen2019toward}
\bibfield{author}{\bibinfo{person}{Bor-Chun Chen} {and} \bibinfo{person}{Andrew
  Kae}.} \bibinfo{year}{2019}\natexlab{}.
\newblock \showarticletitle{Toward realistic image compositing with adversarial
  learning}. In \bibinfo{booktitle}{\emph{Proceedings of the IEEE/CVF
  Conference on Computer Vision and Pattern Recognition}}.
  \bibinfo{pages}{8415--8424}.
\newblock


\bibitem[Chen et~al\mbox{.}(2019)]%
        {chen2019hybrid}
\bibfield{author}{\bibinfo{person}{Kai Chen}, \bibinfo{person}{Jiangmiao Pang},
  \bibinfo{person}{Jiaqi Wang}, \bibinfo{person}{Yu Xiong},
  \bibinfo{person}{Xiaoxiao Li}, \bibinfo{person}{Shuyang Sun},
  \bibinfo{person}{Wansen Feng}, \bibinfo{person}{Ziwei Liu},
  \bibinfo{person}{Jianping Shi}, \bibinfo{person}{Wanli Ouyang},
  {et~al\mbox{.}}} \bibinfo{year}{2019}\natexlab{}.
\newblock \showarticletitle{Hybrid task cascade for instance segmentation}. In
  \bibinfo{booktitle}{\emph{Proceedings of the IEEE/CVF Conference on Computer
  Vision and Pattern Recognition}}. \bibinfo{pages}{4974--4983}.
\newblock


\bibitem[Cheng et~al\mbox{.}(2020)]%
        {cheng2020sequential}
\bibfield{author}{\bibinfo{person}{Yu Cheng}, \bibinfo{person}{Zhe Gan},
  \bibinfo{person}{Yitong Li}, \bibinfo{person}{Jingjing Liu}, {and}
  \bibinfo{person}{Jianfeng Gao}.} \bibinfo{year}{2020}\natexlab{}.
\newblock \showarticletitle{Sequential attention GAN for interactive image
  editing}. In \bibinfo{booktitle}{\emph{Proceedings of the 28th ACM
  International Conference on Multimedia}}. \bibinfo{pages}{4383--4391}.
\newblock


\bibitem[Cohen-Or et~al\mbox{.}(2006)]%
        {cohen2006color}
\bibfield{author}{\bibinfo{person}{Daniel Cohen-Or}, \bibinfo{person}{Olga
  Sorkine}, \bibinfo{person}{Ran Gal}, \bibinfo{person}{Tommer Leyvand}, {and}
  \bibinfo{person}{Ying-Qing Xu}.} \bibinfo{year}{2006}\natexlab{}.
\newblock \showarticletitle{Color harmonization}. In
  \bibinfo{booktitle}{\emph{Proceedings of SIGGRAPH}}.
\newblock


\bibitem[Cong et~al\mbox{.}(2021)]%
        {cong2021bargainnet}
\bibfield{author}{\bibinfo{person}{Wenyan Cong}, \bibinfo{person}{Li Niu},
  \bibinfo{person}{Jianfu Zhang}, \bibinfo{person}{Jing Liang}, {and}
  \bibinfo{person}{Liqing Zhang}.} \bibinfo{year}{2021}\natexlab{}.
\newblock \showarticletitle{Bargainnet: Background-Guided Domain Translation
  for Image Harmonization}. In \bibinfo{booktitle}{\emph{IEEE International
  Conference on Multimedia and Expo}}.
\newblock


\bibitem[Cong et~al\mbox{.}(2020)]%
        {cong2020dovenet}
\bibfield{author}{\bibinfo{person}{Wenyan Cong}, \bibinfo{person}{Jianfu
  Zhang}, \bibinfo{person}{Li Niu}, \bibinfo{person}{Liu Liu},
  \bibinfo{person}{Zhixin Ling}, \bibinfo{person}{Weiyuan Li}, {and}
  \bibinfo{person}{Liqing Zhang}.} \bibinfo{year}{2020}\natexlab{}.
\newblock \showarticletitle{Dovenet: Deep image harmonization via domain
  verification}. In \bibinfo{booktitle}{\emph{Proceedings of the IEEE
  conference on computer vision and pattern recognition}}.
\newblock


\bibitem[Cun and Pun(2020)]%
        {cun2020improving}
\bibfield{author}{\bibinfo{person}{Xiaodong Cun} {and} \bibinfo{person}{Chi-Man
  Pun}.} \bibinfo{year}{2020}\natexlab{}.
\newblock \showarticletitle{Improving the harmony of the composite image by
  spatial-separated attention module}.
\newblock \bibinfo{journal}{\emph{IEEE Transactions on Image Processing}}
  (\bibinfo{year}{2020}).
\newblock


\bibitem[Dosovitskiy et~al\mbox{.}(2021)]%
        {dosovitskiy2021image}
\bibfield{author}{\bibinfo{person}{Alexey Dosovitskiy}, \bibinfo{person}{Lucas
  Beyer}, \bibinfo{person}{Alexander Kolesnikov}, \bibinfo{person}{Dirk
  Weissenborn}, \bibinfo{person}{Xiaohua Zhai}, \bibinfo{person}{Thomas
  Unterthiner}, \bibinfo{person}{Mostafa Dehghani}, \bibinfo{person}{Matthias
  Minderer}, \bibinfo{person}{Georg Heigold}, \bibinfo{person}{Sylvain Gelly},
  {et~al\mbox{.}}} \bibinfo{year}{2021}\natexlab{}.
\newblock \showarticletitle{An image is worth 16x16 words: Transformers for
  image recognition at scale}.
\newblock  (\bibinfo{year}{2021}).
\newblock


\bibitem[Gregor et~al\mbox{.}(2015)]%
        {gregor2015draw}
\bibfield{author}{\bibinfo{person}{Karol Gregor}, \bibinfo{person}{Ivo
  Danihelka}, \bibinfo{person}{Alex Graves}, \bibinfo{person}{Danilo Rezende},
  {and} \bibinfo{person}{Daan Wierstra}.} \bibinfo{year}{2015}\natexlab{}.
\newblock \showarticletitle{Draw: A recurrent neural network for image
  generation}. In \bibinfo{booktitle}{\emph{International Conference on Machine
  Learning}}. PMLR, \bibinfo{pages}{1462--1471}.
\newblock


\bibitem[Guo et~al\mbox{.}(2019)]%
        {guo2019progressive}
\bibfield{author}{\bibinfo{person}{Zongyu Guo}, \bibinfo{person}{Zhibo Chen},
  \bibinfo{person}{Tao Yu}, \bibinfo{person}{Jiale Chen}, {and}
  \bibinfo{person}{Sen Liu}.} \bibinfo{year}{2019}\natexlab{}.
\newblock \showarticletitle{Progressive image inpainting with full-resolution
  residual network}. In \bibinfo{booktitle}{\emph{Proceedings of the 27th acm
  international conference on multimedia}}. \bibinfo{pages}{2496--2504}.
\newblock


\bibitem[Guo et~al\mbox{.}(2021a)]%
        {guo2021image}
\bibfield{author}{\bibinfo{person}{Zonghui Guo}, \bibinfo{person}{Dongsheng
  Guo}, \bibinfo{person}{Haiyong Zheng}, \bibinfo{person}{Zhaorui Gu},
  \bibinfo{person}{Bing Zheng}, {and} \bibinfo{person}{Junyu Dong}.}
  \bibinfo{year}{2021}\natexlab{a}.
\newblock \showarticletitle{Image Harmonization With Transformer}. In
  \bibinfo{booktitle}{\emph{Proceedings of the IEEE/CVF International
  Conference on Computer Vision}}. \bibinfo{pages}{14870--14879}.
\newblock


\bibitem[Guo et~al\mbox{.}(2021b)]%
        {guo2021intrinsic}
\bibfield{author}{\bibinfo{person}{Zonghui Guo}, \bibinfo{person}{Haiyong
  Zheng}, \bibinfo{person}{Yufeng Jiang}, \bibinfo{person}{Zhaorui Gu}, {and}
  \bibinfo{person}{Bing Zheng}.} \bibinfo{year}{2021}\natexlab{b}.
\newblock \showarticletitle{Intrinsic Image Harmonization}. In
  \bibinfo{booktitle}{\emph{Proceedings of the IEEE conference on computer
  vision and pattern recognition}}.
\newblock


\bibitem[Hao et~al\mbox{.}(2020)]%
        {hao2020image}
\bibfield{author}{\bibinfo{person}{Guoqing Hao} {et~al\mbox{.}}}
  \bibinfo{year}{2020}\natexlab{}.
\newblock \showarticletitle{Image Harmonization with Attention-based Deep
  Feature Modulation.}. In \bibinfo{booktitle}{\emph{The British Machine Vision
  Conference}}.
\newblock


\bibitem[Jia et~al\mbox{.}(2006)]%
        {jia2006drag}
\bibfield{author}{\bibinfo{person}{Jiaya Jia}, \bibinfo{person}{Jian Sun},
  \bibinfo{person}{Chi-Keung Tang}, {and} \bibinfo{person}{Heung-Yeung Shum}.}
  \bibinfo{year}{2006}\natexlab{}.
\newblock \showarticletitle{Drag-and-drop pasting}.
\newblock \bibinfo{journal}{\emph{ACM Transactions on graphics}}
  (\bibinfo{year}{2006}).
\newblock


\bibitem[Jiang et~al\mbox{.}(2021)]%
        {jiang2021ssh}
\bibfield{author}{\bibinfo{person}{Yifan Jiang}, \bibinfo{person}{He Zhang},
  \bibinfo{person}{Jianming Zhang}, \bibinfo{person}{Yilin Wang},
  \bibinfo{person}{Zhe Lin}, \bibinfo{person}{Kalyan Sunkavalli},
  \bibinfo{person}{Simon Chen}, \bibinfo{person}{Sohrab Amirghodsi},
  \bibinfo{person}{Sarah Kong}, {and} \bibinfo{person}{Zhangyang Wang}.}
  \bibinfo{year}{2021}\natexlab{}.
\newblock \showarticletitle{SSH: A Self-Supervised Framework for Image
  Harmonization}. In \bibinfo{booktitle}{\emph{Proceedings of the IEEE/CVF
  International Conference on Computer Vision}}. \bibinfo{pages}{4832--4841}.
\newblock


\bibitem[Lalonde and Efros(2007)]%
        {lalonde2007using}
\bibfield{author}{\bibinfo{person}{Jean-Francois Lalonde} {and}
  \bibinfo{person}{Alexei~A Efros}.} \bibinfo{year}{2007}\natexlab{}.
\newblock \showarticletitle{Using color compatibility for assessing image
  realism}. In \bibinfo{booktitle}{\emph{IEEE/CVF International Conference on
  Computer Vision}}.
\newblock


\bibitem[Ling et~al\mbox{.}(2021)]%
        {ling2021region}
\bibfield{author}{\bibinfo{person}{Jun Ling}, \bibinfo{person}{Han Xue},
  \bibinfo{person}{Li Song}, \bibinfo{person}{Rong Xie}, {and}
  \bibinfo{person}{Xiao Gu}.} \bibinfo{year}{2021}\natexlab{}.
\newblock \showarticletitle{Region-aware Adaptive Instance Normalization for
  Image Harmonization}. In \bibinfo{booktitle}{\emph{Proceedings of the IEEE
  conference on computer vision and pattern recognition}}.
\newblock


\bibitem[Liu et~al\mbox{.}(2020)]%
        {liu2020composition}
\bibfield{author}{\bibinfo{person}{Dong Liu}, \bibinfo{person}{Rohit Puri},
  \bibinfo{person}{Nagendra Kamath}, {and} \bibinfo{person}{Subhabrata
  Bhattacharya}.} \bibinfo{year}{2020}\natexlab{}.
\newblock \showarticletitle{Composition-aware image aesthetics assessment}. In
  \bibinfo{booktitle}{\emph{Proceedings of the IEEE/CVF Winter Conference on
  Applications of Computer Vision}}. \bibinfo{pages}{3569--3578}.
\newblock


\bibitem[Liu et~al\mbox{.}(2019)]%
        {liu2019cu}
\bibfield{author}{\bibinfo{person}{Hongying Liu}, \bibinfo{person}{Xiongjie
  Shen}, \bibinfo{person}{Fanhua Shang}, \bibinfo{person}{Feihang Ge}, {and}
  \bibinfo{person}{Fei Wang}.} \bibinfo{year}{2019}\natexlab{}.
\newblock \showarticletitle{CU-Net: Cascaded U-Net with loss weighted sampling
  for brain tumor segmentation}.
\newblock In \bibinfo{booktitle}{\emph{Multimodal Brain Image Analysis and
  Mathematical Foundations of Computational Anatomy}}.
  \bibinfo{publisher}{Springer}, \bibinfo{pages}{102--111}.
\newblock


\bibitem[Peng et~al\mbox{.}(2020)]%
        {peng2020chained}
\bibfield{author}{\bibinfo{person}{Jinlong Peng}, \bibinfo{person}{Changan
  Wang}, \bibinfo{person}{Fangbin Wan}, \bibinfo{person}{Yang Wu},
  \bibinfo{person}{Yabiao Wang}, \bibinfo{person}{Ying Tai},
  \bibinfo{person}{Chengjie Wang}, \bibinfo{person}{Jilin Li},
  \bibinfo{person}{Feiyue Huang}, {and} \bibinfo{person}{Yanwei Fu}.}
  \bibinfo{year}{2020}\natexlab{}.
\newblock \showarticletitle{Chained-tracker: Chaining paired attentive
  regression results for end-to-end joint multiple-object detection and
  tracking}. In \bibinfo{booktitle}{\emph{European Conference on Computer
  Vision}}. Springer, \bibinfo{pages}{145--161}.
\newblock


\bibitem[P{\'e}rez et~al\mbox{.}(2003)]%
        {perez2003poisson}
\bibfield{author}{\bibinfo{person}{Patrick P{\'e}rez}, \bibinfo{person}{Michel
  Gangnet}, {and} \bibinfo{person}{Andrew Blake}.}
  \bibinfo{year}{2003}\natexlab{}.
\newblock \showarticletitle{Poisson image editing}. In
  \bibinfo{booktitle}{\emph{Proceedings of SIGGRAPH}}.
\newblock


\bibitem[Piti{\'e} and Kokaram(2007)]%
        {pitie2007linear}
\bibfield{author}{\bibinfo{person}{Fran{\c{c}}ois Piti{\'e}} {and}
  \bibinfo{person}{Anil Kokaram}.} \bibinfo{year}{2007}\natexlab{}.
\newblock \showarticletitle{The linear monge-kantorovitch linear colour mapping
  for example-based colour transfer}. In \bibinfo{booktitle}{\emph{European
  Conference on Computer Vision}}.
\newblock


\bibitem[Pitie et~al\mbox{.}(2005)]%
        {pitie2005n}
\bibfield{author}{\bibinfo{person}{Francois Pitie}, \bibinfo{person}{Anil~C
  Kokaram}, {and} \bibinfo{person}{Rozenn Dahyot}.}
  \bibinfo{year}{2005}\natexlab{}.
\newblock \showarticletitle{N-dimensional probability density function transfer
  and its application to color transfer}. In \bibinfo{booktitle}{\emph{IEEE/CVF
  International Conference on Computer Vision}}.
\newblock


\bibitem[Qiu et~al\mbox{.}(2020)]%
        {qiu2020semanticadv}
\bibfield{author}{\bibinfo{person}{Haonan Qiu}, \bibinfo{person}{Chaowei Xiao},
  \bibinfo{person}{Lei Yang}, \bibinfo{person}{Xinchen Yan},
  \bibinfo{person}{Honglak Lee}, {and} \bibinfo{person}{Bo Li}.}
  \bibinfo{year}{2020}\natexlab{}.
\newblock \showarticletitle{SemanticAdv: Generating Adversarial Examples via
  Attribute-conditioned Image Editing}. In \bibinfo{booktitle}{\emph{European
  Conference on Computer Vision}}. Springer, \bibinfo{pages}{19--37}.
\newblock


\bibitem[Rodriguez and Laio(2014)]%
        {rodriguez2014clustering}
\bibfield{author}{\bibinfo{person}{Alex Rodriguez} {and}
  \bibinfo{person}{Alessandro Laio}.} \bibinfo{year}{2014}\natexlab{}.
\newblock \showarticletitle{Clustering by fast search and find of density
  peaks}.
\newblock \bibinfo{journal}{\emph{science}} \bibinfo{volume}{344},
  \bibinfo{number}{6191} (\bibinfo{year}{2014}), \bibinfo{pages}{1492--1496}.
\newblock


\bibitem[Ronneberger et~al\mbox{.}(2015)]%
        {ronneberger2015u}
\bibfield{author}{\bibinfo{person}{Olaf Ronneberger}, \bibinfo{person}{Philipp
  Fischer}, {and} \bibinfo{person}{Thomas Brox}.}
  \bibinfo{year}{2015}\natexlab{}.
\newblock \showarticletitle{U-net: Convolutional networks for biomedical image
  segmentation}. In \bibinfo{booktitle}{\emph{International Conference on
  Medical Image Computing and Computer Assisted Intervention}}.
\newblock


\bibitem[Sofiiuk et~al\mbox{.}(2021)]%
        {sofiiuk2021foreground}
\bibfield{author}{\bibinfo{person}{Konstantin Sofiiuk}, \bibinfo{person}{Polina
  Popenova}, {and} \bibinfo{person}{Anton Konushin}.}
  \bibinfo{year}{2021}\natexlab{}.
\newblock \showarticletitle{Foreground-aware Semantic Representations for Image
  Harmonization}. In \bibinfo{booktitle}{\emph{Proceedings of the IEEE/CVF
  Winter Conference on Applications of Computer Vision}}.
  \bibinfo{pages}{1620--1629}.
\newblock


\bibitem[Song et~al\mbox{.}(2020)]%
        {song2020illumination}
\bibfield{author}{\bibinfo{person}{Shuangbing Song}, \bibinfo{person}{Fan
  Zhong}, \bibinfo{person}{Xueying Qin}, {and} \bibinfo{person}{Changhe Tu}.}
  \bibinfo{year}{2020}\natexlab{}.
\newblock \showarticletitle{Illumination harmonization with gray mean scale}.
  In \bibinfo{booktitle}{\emph{Computer Graphics International Conference}}.
  Springer, \bibinfo{pages}{193--205}.
\newblock


\bibitem[Sunkavalli et~al\mbox{.}(2010)]%
        {sunkavalli2010multi}
\bibfield{author}{\bibinfo{person}{Kalyan Sunkavalli}, \bibinfo{person}{Micah~K
  Johnson}, \bibinfo{person}{Wojciech Matusik}, {and}
  \bibinfo{person}{Hanspeter Pfister}.} \bibinfo{year}{2010}\natexlab{}.
\newblock \showarticletitle{Multi-scale image harmonization}.
\newblock \bibinfo{journal}{\emph{ACM Transactions on Graphics}}
  (\bibinfo{year}{2010}).
\newblock


\bibitem[Tao et~al\mbox{.}(2013)]%
        {tao2013error}
\bibfield{author}{\bibinfo{person}{Michael~W Tao}, \bibinfo{person}{Micah~K
  Johnson}, {and} \bibinfo{person}{Sylvain Paris}.}
  \bibinfo{year}{2013}\natexlab{}.
\newblock \showarticletitle{Error-tolerant image compositing}.
\newblock \bibinfo{journal}{\emph{International journal of computer vision}}
  (\bibinfo{year}{2013}).
\newblock


\bibitem[Tsai et~al\mbox{.}(2017)]%
        {tsai2017deep}
\bibfield{author}{\bibinfo{person}{Yi-Hsuan Tsai}, \bibinfo{person}{Xiaohui
  Shen}, \bibinfo{person}{Zhe Lin}, \bibinfo{person}{Kalyan Sunkavalli},
  \bibinfo{person}{Xin Lu}, {and} \bibinfo{person}{Ming-Hsuan Yang}.}
  \bibinfo{year}{2017}\natexlab{}.
\newblock \showarticletitle{Deep image harmonization}. In
  \bibinfo{booktitle}{\emph{Proceedings of the IEEE conference on computer
  vision and pattern recognition}}.
\newblock


\bibitem[Van~den Oord et~al\mbox{.}(2016)]%
        {van2016conditional}
\bibfield{author}{\bibinfo{person}{Aaron Van~den Oord}, \bibinfo{person}{Nal
  Kalchbrenner}, \bibinfo{person}{Lasse Espeholt}, \bibinfo{person}{Oriol
  Vinyals}, \bibinfo{person}{Alex Graves}, {et~al\mbox{.}}}
  \bibinfo{year}{2016}\natexlab{}.
\newblock \showarticletitle{Conditional image generation with pixelcnn
  decoders}.
\newblock \bibinfo{journal}{\emph{Advances in neural information processing
  systems}}  \bibinfo{volume}{29} (\bibinfo{year}{2016}).
\newblock


\bibitem[Vaswani et~al\mbox{.}(2017)]%
        {vaswani2017attention}
\bibfield{author}{\bibinfo{person}{Ashish Vaswani}, \bibinfo{person}{Noam
  Shazeer}, \bibinfo{person}{Niki Parmar}, \bibinfo{person}{Jakob Uszkoreit},
  \bibinfo{person}{Llion Jones}, \bibinfo{person}{Aidan~N Gomez},
  \bibinfo{person}{{\L}ukasz Kaiser}, {and} \bibinfo{person}{Illia
  Polosukhin}.} \bibinfo{year}{2017}\natexlab{}.
\newblock \showarticletitle{Attention is all you need}.
\newblock \bibinfo{journal}{\emph{Advances in neural information processing
  systems}}  \bibinfo{volume}{30} (\bibinfo{year}{2017}).
\newblock


\bibitem[Wang et~al\mbox{.}(2020)]%
        {wang2020image}
\bibfield{author}{\bibinfo{person}{Jin Wang}, \bibinfo{person}{Chen Wang},
  \bibinfo{person}{Qingming Huang}, \bibinfo{person}{Yunhui Shi},
  \bibinfo{person}{Jian-Feng Cai}, \bibinfo{person}{Qing Zhu}, {and}
  \bibinfo{person}{Baocai Yin}.} \bibinfo{year}{2020}\natexlab{}.
\newblock \showarticletitle{Image inpainting based on multi-frequency
  probabilistic inference model}. In \bibinfo{booktitle}{\emph{Proceedings of
  the 28th ACM International Conference on Multimedia}}. \bibinfo{pages}{1--9}.
\newblock


\bibitem[Wu et~al\mbox{.}(2020)]%
        {wu2020cascade}
\bibfield{author}{\bibinfo{person}{Rongliang Wu}, \bibinfo{person}{Gongjie
  Zhang}, \bibinfo{person}{Shijian Lu}, {and} \bibinfo{person}{Tao Chen}.}
  \bibinfo{year}{2020}\natexlab{}.
\newblock \showarticletitle{Cascade ef-gan: Progressive facial expression
  editing with local focuses}. In \bibinfo{booktitle}{\emph{Proceedings of the
  IEEE/CVF Conference on Computer Vision and Pattern Recognition}}.
  \bibinfo{pages}{5021--5030}.
\newblock


\bibitem[Xue et~al\mbox{.}(2012)]%
        {xue2012understanding}
\bibfield{author}{\bibinfo{person}{Su Xue}, \bibinfo{person}{Aseem Agarwala},
  \bibinfo{person}{Julie Dorsey}, {and} \bibinfo{person}{Holly Rushmeier}.}
  \bibinfo{year}{2012}\natexlab{}.
\newblock \showarticletitle{Understanding and improving the realism of image
  composites}.
\newblock \bibinfo{journal}{\emph{ACM Transactions on graphics}}
  (\bibinfo{year}{2012}).
\newblock


\bibitem[Zhan et~al\mbox{.}(2020)]%
        {zhan2020adversarial}
\bibfield{author}{\bibinfo{person}{Fangneng Zhan}, \bibinfo{person}{Shijian
  Lu}, \bibinfo{person}{Changgong Zhang}, \bibinfo{person}{Feiying Ma}, {and}
  \bibinfo{person}{Xuansong Xie}.} \bibinfo{year}{2020}\natexlab{}.
\newblock \showarticletitle{Adversarial image composition with auxiliary
  illumination}. In \bibinfo{booktitle}{\emph{Proceedings of the Asian
  Conference on Computer Vision}}.
\newblock


\bibitem[Zhu et~al\mbox{.}(2015)]%
        {zhu2015learning}
\bibfield{author}{\bibinfo{person}{Jun-Yan Zhu}, \bibinfo{person}{Philipp
  Krahenbuhl}, \bibinfo{person}{Eli Shechtman}, {and} \bibinfo{person}{Alexei~A
  Efros}.} \bibinfo{year}{2015}\natexlab{}.
\newblock \showarticletitle{Learning a discriminative model for the perception
  of realism in composite images}. In \bibinfo{booktitle}{\emph{IEEE/CVF
  International Conference on Computer Vision}}.
\newblock


\end{thebibliography}

\end{document}